\documentclass{article}

\usepackage{times}
\usepackage{graphicx} 
\usepackage{subfigure}

\usepackage{natbib}

\usepackage{algorithm}
\usepackage{algorithmic}

\usepackage{hyperref}

\usepackage{amsmath}

\usepackage[accepted]{icml2017}

\icmltitlerunning{Learning What Data to Learn}

\begin{document}

\twocolumn[
\icmltitle{Learning What Data to Learn}



\icmlsetsymbol{equal}{*}
\begin{icmlauthorlist}
\icmlauthor{Yang Fan}{ustc}
\icmlauthor{Fei Tian}{microsoft}
\icmlauthor{Tao Qin}{microsoft}
\icmlauthor{Jiang Bian}{microsoft}
\icmlauthor{Tie-Yan Liu}{microsoft}
\end{icmlauthorlist}

\icmlaffiliation{ustc}{University of Science and Technology of China, Hefei, China}
\icmlaffiliation{microsoft}{Microsoft Research, Beijing, China}

\icmlcorrespondingauthor{Fei Tian}{fetia@microsoft.com}
\icmlcorrespondingauthor{Yang Fan}{fyabc@mail.ustc.edu.cn}


\vskip 0.3in
]



\printAffiliationsAndNotice{} 

\begin{abstract}

Machine learning is essentially the sciences of playing with data. An adaptive data selection strategy, enabling to dynamically choose different data at various training stages, can reach a more effective model in a more efficient way. In this paper, we propose a deep reinforcement learning framework, which we call \emph{\textbf{N}eural \textbf{D}ata \textbf{F}ilter} (\textbf{NDF}), to explore automatic and adaptive data selection in the training process. In particular, NDF takes advantage of a deep neural network to adaptively select and filter important data instances from a sequential stream of training data, such that the future accumulative reward (e.g., the convergence speed) is maximized. In contrast to previous studies in data selection that is mainly based on heuristic strategies, NDF is quite generic and thus can be widely suitable for many machine learning tasks. Taking neural network training with stochastic gradient descent (SGD) as an example, comprehensive experiments with respect to various neural network modeling (e.g., multi-layer perceptron networks, convolutional neural networks and recurrent neural networks) and several applications (e.g., image classification and text understanding) demonstrate that NDF powered SGD can achieve comparable accuracy with standard SGD process by using less data and fewer iterations.

\end{abstract}

\section{Introduction}

Training data plays a critical role in machine learning. The data selection strategy along the training process could significantly impact the performance of the learned model. For example, an appropriate strategy of removing redundant data can lead to a better model by using less computational efforts. For another example, previous studies on curriculum learning ~\citep{CL} and self-paced learning~\citep{SPL} reveal some principles of  how tailoring data based on its `hardness' can favor the model training; that is, \emph{easy} data instances are important at the early age of model training, while at later age, \emph{harder} training examples tend to be more effective to improve the model, since \emph{easy} ones bring minor changes.

These facts reveal that how to feed training samples into a machine learning system is nontrivial and feeding data in a totally random order is not always a good choice. To explore a better data selection strategy for training, previous works including curriculum learning (CL) and self-paced learning (SPL) adopt simple heuristic rules, such as shuffling the sequence length to train language model ~\citep{CL}, or abandoning training instances whose loss values are larger than a human-defined threshold ~\citep{SPL,SPL4MM}. Such human-defined rules are a little restricted to certain tasks and cannot be generalized to broader learning scenarios, since different learning tasks may yield different optimal data selection rules, and even one learning task may need data with various properties to optimize at different training stages. Therefore it remains an open problem \emph{how to automatically and dynamically allocate appropriate training data at different stages of machine learning}?

To find a solution to the above problem, we design two-fold intuitive principles: on one hand, the data selection strategy should be \emph{general} enougg, such that it can be naturally applied to different learning scenarios without further particularly human-designed efforts; on the other hand, the strategy should be \emph{forward-looking}, in that its choice at every step along the training leads to better long-term reward, rather than temporarily fitting to current stage. 

Following these principles, we propose a new data selection framework, based on deep reinforcement learning (DRL). In this framework, the DRL based data selection model acts at a \emph{teacher} while the training process of the target model is a \emph{student}, and the \emph{teacher} is responsible for providing the \emph{student} with the appropriate training data. This \emph{teacher}-\emph{student} framework can not only make it possible to models the long-term reward along with training process, but also be well generalized to most machine learning scenarios, since reinforcement learning approach can model data selection mechanism as a parametric policy that can be adaptive, work on flexible state spaces that cover any signals used in previous work, and be automatically optimized in an end-to-end way.  

To better elaborate our proposal, we focus on applying DRL to mini-batch stochastic gradient descent (SGD) that is widely used to optimize a machine learning model in the training process. Mini-batch SGD is a sequential process, in which mini-batches of data $D=\{D_1,\cdots D_t,\dots,D_T\}$ arrive sequentially in a random order. Here $D_t=(d_1,\cdots,d_M)$ is the mini-batch of data arriving at the $t$-th time step and consisting of $M$ training instances. Given $D_t$, the loss and gradient w.r.t. the current model $\mathcal{W}_t$ are respectively denoted as $L_t=\frac{1}{M}\sum_{m=1}^Ml(d_m)$ and $g_t = \frac{\partial L_t}{\partial \mathcal{W}_t}$. Then, mini-batch SGD updates the model as follows:

\begin{equation}
\label{eqn:nn_weight_trans}
\mathcal{W}_{t+1}=\mathcal{W}_t - \eta_tg_t.
\end{equation}Here $l(\cdot)$ is the loss function and $\eta_t$ is the learning rate at $t$-th step.

\begin{figure}
\centering
\includegraphics[width=\columnwidth]{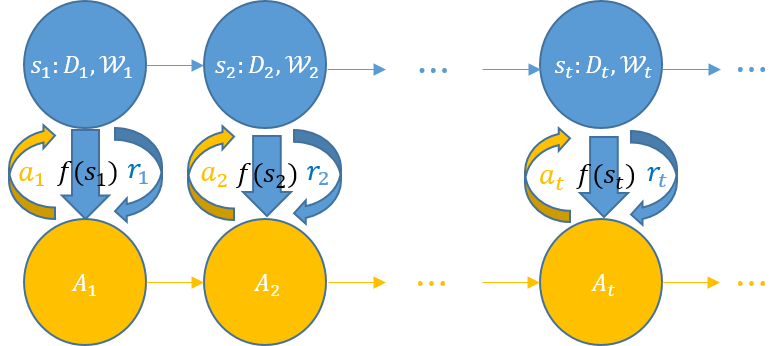}
\caption{The structure of SGD accompanied with NDF. Blue part refers to SGD training process and yellow part is NDF.}
\label{fig:ndf}
\vspace{-3mm}
\end{figure}

By assuming the use of mini-batch SGD, our proposed method aims at dynamically determining which instances in mini-batch $D_t$ are used for training and which are abandoned, after receiving $D_t$ from $M$ training instances,. Specifically, we use deep reinforcement learning to determine whether/how to filter the given mini-batch of training data, which we call \textbf{Neural Data Filter} (NDF). In NDF, as illustrated in Figure \ref{fig:ndf}, the SGD training for the base machine learning model (i.e., the trainee) is casted into a Markov Decision Process (MDP) ~\citep{RLSutton}. In such an MDP, a state (namely $s_1,\cdots,s_t,\cdots$), characterizing current state of training process, is composed of two parts: the mini-batch of arrived data and the current parameters of the trainee , i.e., $s_t=\{D_t,\mathcal{W}_t\}$. In each time step $t$, NDF receives a representation $f(s_t)$ for the current state from SGD, and outputs the action $a_t$ specifying which instances in $D_t$ will be filtered according to its policy $A_t$. Afterwards, the remaining data determined by $a_t$ will be used by SGD to update the trainee's state and generate a reward $r_t$ (such as validation accuracy), which will be conversely leveraged by NDF as the feedback for updating its own policy.

We apply NDF to training various types of neural networks, including MLP, CNN and RNN, with training data from different domains including image and text. Experimental studies demonstrates faster convergence speed caused by NDF over baselines. Further analysis shows that :1) a well-designed data selection policy benefits deep neural network training, which has not been fully explored in the community. 2) The automatically learnt data selection policy based on reinforcement learning is quite general, surpassing human designed efforts for each task.

The rest of the paper is organized as follows. In Section \ref{sec:NDF}, in the context of mini-batch SGD algorithms, we introduce the details of Neural Data Filter (NDF), including the MDP language to perform training data filtration, and the policy gradient algorithms to learn NDF. Then in Section ~\ref{sec:Exp}, taking deep neural network training as example, we empirically verify the effectiveness of NDF. We discuss related work in Section \ref{sec:Related} and conclude the paper in the last section.

\section{Neural Data Filter}
\label{sec:NDF}

In this section, by assuming training machine learning models with mini-batch SGD, we introduce the mathematical details of Neural Data Filter (NDF). As a summary, NDF aims to filter certain amount of training data within a mini-batch, such that only high-quality training data is remained and better convergence speed for SGD training is achieved. To achieve that, as introduced in Section~\ref{sec:Intro} and Figure \ref{fig:ndf}, we cast SGD training as a Markov Decision Process (MDP), termed as \textbf{SGD-MDP}.

\textbf{SGD-MDP}: As traditional MDP, SGD-MDP is composed of the tuple $<s,a,\mathcal{P},r, \gamma>$, illustrated as:
\begin{itemize}
\item $s$ is the state, corresponding to the mini-batch data arrived and current state of machine learning model (i.e., the trainee): $s_t=(D_t, \mathcal{W}_t)$.
\item $a$ represents the space of actions. For data filtration task, we have $a =\{a_m\}_{m=1}^M\in \{0,1\}^M$, where $M$ is the batch size and $a_m\in\{0,1\}$ denotes whether to filter the $m$-th data instance in $D_t$ or not\footnote{We consider data instances within the same mini-batch are independent with each other, and therefore for statement simplicity, when the context is clear, $a$ will be used to denote the remain/filter decision for single data instance, i.e., $a\in\{1,0\}$. Similarly, the notation $s$ will sometimes represent the state for only one training instance.}. Those filtered instances will have no effects to base model training.
\item $\mathcal{P}^a_{ss'}=P(s'|s,a)$ is the state transition probability, determined by two factors: 1) The uniform distribution of sequentially arrived training batch data; 2) The optimization process specified by Gradient Descent principle (c.f. Equation \ref{eqn:nn_weight_trans}). The randomness comes from stochastic factors in training, such as dropout ~\citep{Dropout}.
\item $r=r(s,a)$ is the reward, set to be any signal indicating how well the training goes, such as validation accuracy, or the loss gap for current mini-batch data before/after model update.
\item Furthermore future reward $r$ is discounted by a discounting factor $\gamma\in[0,1]$ into the cumulative reward.
\end{itemize}

NDF samples the action $a$ by its policy function $A=P_\Theta(a|s)$ with parameters $\Theta$ to be learnt. The policy $A$ can be any binary classification model, such as logistic regression and deep neural network. For example, $A(s,a;\Theta)=P_\Theta(a|s)= a\sigma(\theta f(s)+b)+ (1-a)(1-\sigma(\theta f(s)+b))$, where $\sigma(\cdot)$ is the sigmoid function, $\Theta=\{\theta, b\}$ and $f(s)$ is the feature vector to effectively represent state $s$, discussed as below.



\textbf{State Features}: The aim of designing state feature vector $f(s)$ is to effectively and efficiently represent SGD-MDP state. Since state $s$ includes both arrived training data and current base model state, we adopt three categories features to compose $f(s)$:
\begin{itemize}
\item Data features, contain information for data instance, such as its label category (we use \emph{$1$ of $|Y|$} representations), (for texts) the length of sentence, linguistic features for text segments~\citep{CL4wordembedding}, or (for images) gradients histogram features ~\citep{histoFea}. Such data features are commonly used in curriculum learning ~\citep{CL,CL4wordembedding}.
\item Base model features, include the signals reflecting how \emph{well} current machine learning model is trained. We collect several simple features, such as passed mini-batch number (i.e., iteration), the average historical training loss and historical validation accuracy. They are proven to be effective enough to represent current model status.
\item Features to represent the combination of both data and model. By using these features, we target to represent how \emph{important} the arrived training data is for current model. We mainly use three parts of such signals in our classification tasks: 1) the predicted probabilities of each class; 2) the loss value on that data, which appears frequently in self-paced learning algorithms~\citep{SPL,SPL4MM,CL4QA}; 3) the margin value \footnote{The margin for a training instance $(x,y)$ is defined as $P(y|x)-\max_{y'\neq y}P(y'|x)$~\citep{cortes2013multi}}.
\end{itemize}
The state features $f(s)$ are computed after the arrival of each mini-batch of training data.

The whole process for training with NDF is listed in Algorithm \ref{alg:whole}. In particular, we take the similar generalization  framework proposed in ~\citep{L2R}, in which we randomly sample a subset of training data to train the policy of NDF (Step 1 and 2) with policy gradient method, and apply the data filtration model to the training process on the whole dataset (Step 3). The detailed algorithm to train NDF policy will be introduced in the next subsection.

\begin{algorithm}
\caption{SGD Training with Neural Data Filter.}\label{alg:whole}
\begin{algorithmic}
\STATE \textbf{Input}: Training data $D$.
\STATE 1. Randomly sample a subset of NDF training data $D'$ from $D$.
\STATE 2. Optimize NDF policy network $A(s;\Theta)$ based on $D'$ by policy gradient (details in Algorithm \ref{alg:REINFORCE}).
\STATE 3. Apply $A(s;\Theta)$ to full dataset $D$ to train the base machine learning model by SGD.
\STATE \textbf{Output}: The base machine learning model.
\end{algorithmic}
\end{algorithm}

\subsection{Training Algorithm for NDF Policy}

To obtain the optimal data filtration policy, we aim to optimize the following expected reward:

\begin{equation}
\label{eqn:exp_reward}
J(\Theta)=E_{P_\Theta(a|s)}[R(s,a)],
\end{equation} where $R(s,a)$ is the state-action value function and $\Theta$ parameterizes the policy. Since $R(s,a)$ is non-differentiable w.r.t. $\Theta$, we use REINFORCE ~\citep{REINFORCE}, a Monte-Carlo policy gradient algorithm to optimize the above quantity in Equation (\ref{eqn:exp_reward}):

\begin{equation}
\nabla_\Theta=\sum_{t=1}^TE_{P_\Theta(a_{1:T}|s)}[ \nabla_\Theta \log P(a_t|s_t)R(s_t,a_t)],
\end{equation}

which is empirically estimated as:

\begin{equation}
\label{eqn:grad_REINFORCE}
\sum_{t=1}^T\nabla_\Theta\log P(a_t|s_t)v_t.
\end{equation}

In our scenario, $v_t$ is the sampled estimation of $R(s_t,a_t)$ from one episode execution of data filtration policy $P_\Theta(a|s)$: $v_t=r_t+\gamma r_{t+1}+\cdots+\gamma^{T-t}r_T$, where $r_t$ is the sampled reward (e.g., accuracy on a held-out validation set in $D'$) at time-step $t$, and $\gamma\in[0,1]$ is a discount factor. To further reduce the high variance of the gradient estimation in Equation (\ref{eqn:grad_REINFORCE}), we use some variance reduction technique such as substracting reward baseline function~\cite{Rew_Base} and the details will be shown in subsection~\ref{subsec:exp_setup}.



The flow of training NDF policy is given in Algorithm~\ref{alg:REINFORCE}.

\begin{algorithm}[ht]
\caption{Train NDF policy.}\label{alg:REINFORCE}
\begin{algorithmic}
\STATE \textbf{Input}: Training data $D'$. Episode number $L$. Mini-batch size $M$. Discount factor $\gamma\in[0,1]$.
\STATE Randomly split $D'$ into two disjoint subsets: $D'_{train}$ and $D'_{dev}$.
\STATE Initialize NDF data filtration policy $A(s,a;\Theta)$, i.e., $P_\Theta(a|s)$.
\FOR {each episode $l=1,2,\cdots,L$}
\STATE Initialize the base machine learning model.
\STATE Shuffle $D'_{train}$ to get the mini-batches sequence $\{D_1,D_2,\cdots\}$.
\STATE $T=0$.
\WHILE {stopping criteria is not met}
\STATE $T=T+1$.
\STATE Sample data filtration action for each data instance in $D_T=\{d_1,\cdots,d_M\}$: $a=\{a_m\}_{m=1}^M$, $a_m\propto P_\Theta(a|s_m)$, $s_m$ is the state corresponding to $d_m$.
\STATE Update base machine learning model by Gradient Descent based on the selected data in $D_T$.
\STATE Receive reward $r_T$ computed on $D'_{dev}$.
\ENDWHILE

\FOR {$t=1,\cdots,T$}
\STATE Compute cumulative reward $v_t=r_t+\gamma r_{t+1}+\cdots+\gamma^{T-t}r_T$.
\begin{equation}
\label{eqn:ndf_reinforce}
\Theta\leftarrow \Theta +\alpha v_t\sum_m\frac{\partial\log P_\Theta(a_m|s_m)}{\partial \Theta}
\end{equation}
\ENDFOR
\ENDFOR
\STATE \textbf{Output}: The NDF policy $A(s,a;\Theta)$. 
\end{algorithmic}
\end{algorithm}

\section{Experiments}
\label{sec:Exp}

In this section, taking neural networks training as an example, we demonstrate NDF improves SGD's convergence performance by a large margin. The experiments are conducted with three most commonly used neural networks: multi-layer perceptron (MLP), convolutional neural networks (CNN) and recurrent neural networks (RNN), on both image and text classification tasks.

\subsection{Experiments Setup}
\label{subsec:exp_setup}

Different data filtration strategies we applied to SGD training include:

\begin{itemize}
\item \textbf{Unfiltered SGD}. The SGD training algorithm without any data filtration. Here rather than vanilla SGD (c.f. Equation (\ref{eqn:nn_weight_trans})), we use its advanced variants such as Adadelta ~\citep{Adadelta} or Momentum-SGD ~\citep{momentum_sgd} to perform base model training in each task. 

\item \textbf{Self-Paced Learning (SPL)}~\citep{SPL}. It refers to filtering training data by its \emph{hardness}, as reflected by loss value. Mathematically speaking, those training data $d$ satisfying $l(d)> \eta$ will be filtered out, where the threshold $\eta$ grows from smaller to larger during the training process.

In our implementation, to improve the robustness of SPL, following the widely used trick in common SPL implementation ~\citep{SPLDiversity}, we filter training data using its loss rank in one mini-batch, rather than the absolute loss value \footnote{As we empirically tested, filtering by loss value will lead to quite slow and unstable convergence in model training.}. That is to say, we filter data instances with top $K$ largest training loss values within a $M$-sized mini-batch, where $K$ linearly drops from $M-1$ to $0$ during training.

\item \textbf{NDF}. SGD training with data filtration mechanism learnt by NDF, as shown in Algorithm \ref{alg:REINFORCE}.

The state features are constructed according to the principles described in \textbf{State Features} of Section \ref{sec:NDF}. In our experiments, we define the reward in the following way: we set an accuracy threshold $\tau\in[0,1]$ and for each episode $l$, record the first mini-batch index $i_\tau$ in which the accuracy on held-out dev set $D'_{dev}$ exceeds $\tau$, then set the reward as $r_l=-\log(\tau/{T'})$,  where $T'$ is a pre-defined maximum iteration number. Note that here only terminal reward exists. There are many other ways to define the reward, and we will explore them in our future work.

We use a three-layer neural network as the data filtration policy function. All the weight values in this network are uniformly initialized between $(-0.01,0.01)$. The bias terms are all set as $0$ except for the bias in the last-layer which is initialized as $2$, with the goal of not filtering too much data in the early age. Adam ~\citep{Adam} is leveraged to optimize the policy. The policy that achieves the best terminal reward in all episodes is applied as the final policy to the full dataset (c.f., Line 3 in Algorithm~\ref{alg:whole}). To reduce estimation variance, a moving average of the historical reward values in previous episodes is set as a reward baseline for the current episode ~\cite{Rew_Base}. That is, switching Eqn. \ref{eqn:ndf_reinforce} to:

\begin{equation}
\Theta\leftarrow \Theta +\alpha (r_t-b_l)\sum_m\frac{\partial\log P_\Theta(a|s_m)}{\partial \Theta}.
\end{equation}with the reward baseline $b_l$ for episode $l,  l=1,2,\cdots, L$, computed as $b_l=0.8 b_{l-1} + 0.2r_l, b_0 = 0$, and $v_t$ is in fact computed as $v_t=r_l$.


\item \textbf{RandDrop.} To conduct more comprehensive comparison, for NDF, we record the ratio of filtered data instances per epoch, and then randomly filter data in each epoch according to the logged ratio. In this way we form one more baseline, referred to as RandDrop.

\end{itemize}

For all strategies other than Unfiltered SGD, we make sure that \emph{the base neural network model will not be updated until $M$ un-trained, yet selected data instances are accumulated}. In this way, we guarantee that when updating neural network parameters, the batch sizes are the same for every strategy (i.e., $M$), and thus the convergence speed is only determined by the quality of selected data,  not by different model update frequencies since data filtration within a mini-batch will (otherwise) lead to smaller batch size. The model is implemented with Theano~\cite{Theano} and run on one Tesla K40 GPU for each training/testing process.

For each data filtration strategy in every task, we report the test accuracy with respect to the number of effective training instances. To demonstrate the robustness of NDF, we set different hyper-parameters, both for NDF and SPL, and then plot the curve for each hyper-parameter configuration.  Concretely speaking, for NDF, we vary the validation threshold $\tau$ in reward computation. For SPL, we test different speeds to embrace all the training data during training process. Such a speed is characterized by a pre-defined epoch number $S$, which means all the training data will gradually be included (i.e., $K$ linearly drops from $M-1$ to $0$) among the first $S$ epochs.  All the experimental curves reported below are the average results of $5$ repeated runs.

\begin{figure}[htbp]
\centering
\includegraphics[width=1\columnwidth,height=5.4cm]{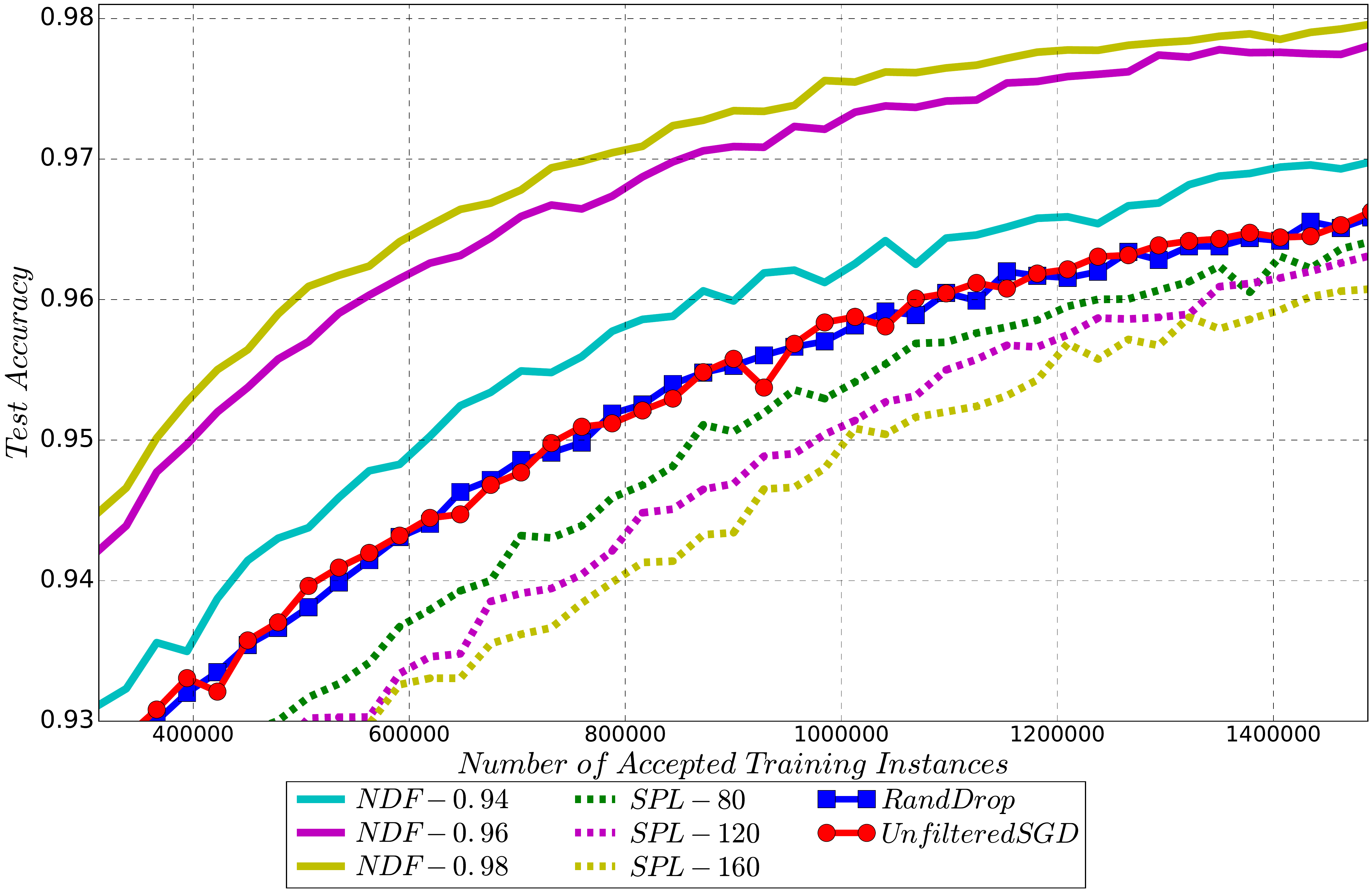}
\caption{Test accuracy curves of different data filtration strategies on MNIST dataset.  Different hyper-parameter settings are included: NDF with validation accuracy threshold set as $\tau=0.94$, $0.96$ and $0.98$, SPL with $S$ respectively configured as $80,120,160$. RandDrop uses the filtered data information output by NDF with $\tau=0.98$. The $x$-axis records the number of effective training instances. }
\label{fig:mnist}
\vspace{-3mm}
\end{figure}

\subsection{MLP for MNIST}

We first test different data filtration strategies for multilayer perceptron network training on image recognition task. The dataset we used is MNIST, which consists of $60k$ training and $10k$ testing images of handwritten digits from 10 categories (i.e., 0, $\cdots$, 9). Momentum-SGD with mini-batch size as $20$ is used to perform MLP model training.

A three-layer feedforward neural network with layer size $784\times 500 \times 10$ and cross-entropy loss is used to classify the MNIST dataset. $tanh$ acts as the activation function for the hidden layer.  The subset $D'$ (c.f.,  Algorithm~\ref{alg:whole}) contains $50k$ randomly selected images from the whole training set and $5k$ instances in $D'$ serves as the held-out validation set $D'_{dev}$.  We train the policy network for $L=500$ episodes and we control training in each episode by early stopping based on validation set accuracy.  NDF leverages a three-layer neural network with model size $25\times 12\times 1$ as policy network, where the first layer node number $25$ is the dimension of state features $f_s$.  $tanh$ function is the activation function for the middle layer.  

The accuracy curves of different data filtration strategies on the test set are plotted in Figure \ref{fig:mnist}.  From Figure \ref{fig:mnist} we can observe that NDF achieves the best convergence speed, significantly better than other policies.  In particular, with $\tau$ set as $0.98$ , to achieve a fairly good classification accuracy e.g,  $0.97$, SGD with NDF uses much less training data (about $750k$) than SGD without any data selection mechanism (roughly $1.5M$). SPL does not select important data in model training, as reflected by the curves in scattered dots. 

\begin{figure}[htbp]
\centering
\includegraphics[width=1\columnwidth,height=4.3cm]{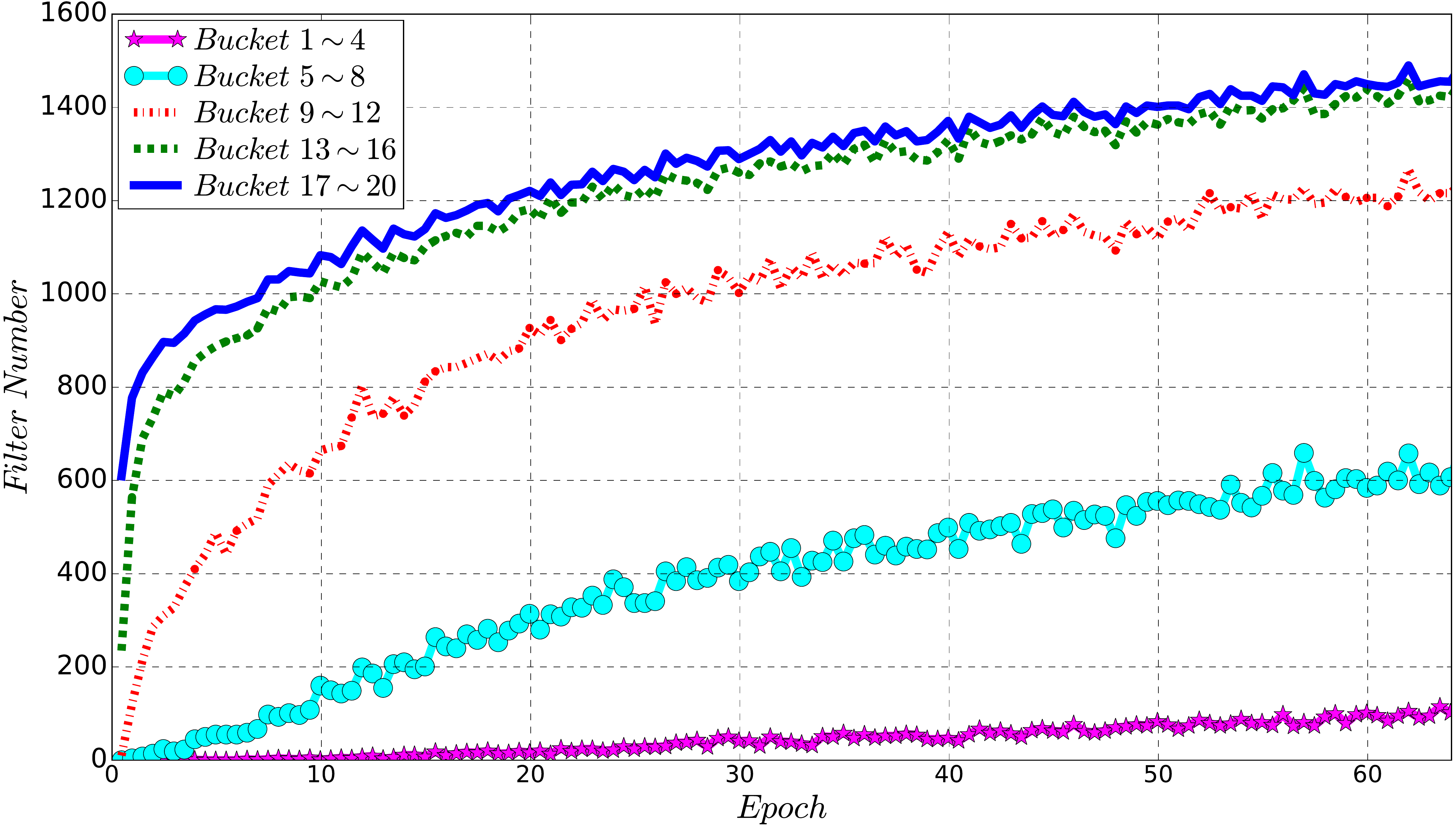}
\caption{The filtered data numbers by NDF in each epoch of MNIST training. Different curves denote the number of filtered data corresponding to different \emph{hardness} levels, as indicated by the ranks of loss on that filtered data instance within its mini-batch. Concretely speaking, we category all the rank values $\{1,2,\cdots, 20\}$, where $20$ is the number of training instances in each mini-batch, into five buckets. Bucket $1\sim 4	$ denotes the \emph{hardest} data instances whose loss values are largest (ranked top  $1$ to $4$) among each mini-batch, while Bucket $17\sim 20$ is the \emph{easiest} among each batch, with smallest loss values.}
\label{fig:mnist_dp}
\vspace{-3mm}
\end{figure}

To better understand the behaviors of NDF, in Figure \ref{fig:mnist_dp}, we record the number of instances filtered by NDF in each epoch, and use five curves to denote the number of filtered instances corresponding to difference \emph{hardness}  levels,  which are measured by its rank of loss values among all the data in its mini-batch. From this figure it is clearly observed that the data selection strategy is quite different from SPL: First, as the training goes on, more and more data will be filtered, which is opposite to that of SPL. Second,  \emph{hard} data (with \emph{hardest} category shown in the purple curve) tend to be selected for training, while \emph{easy} ones (with \emph{easiest} category shown in the blue line) will probably be filtered. We believe such a result well demonstrates that training MLP on MNIST favors the critical data which brings fairly larger effects to model training, whereas those less informative data instances, with smaller loss values,  are comparatively redundant and negligible.

\subsection{CNN for Cifar10}

In this subsection, we conduct experiments on a larger vision dataset than MNIST, with more powerful classification model than MLP. Specifically, we use \emph{CIFAR-10}~\citep{Cifar10}, a widely used dataset for image classification, which contains $60k$ RGB images of size $32\times 32$ categorized into $10$ classes. The dataset is partitioned into a training set with $50k$ images and a test set with $10k$ images. Furthermore, data augmentation is applied to every training image, with padding 4 pixels to each side and randomly sampling a $32\times 32$ crop. ResNet~\citep{ResNet}, a well-known effective CNN model for image recognition, is adopted to perform classification on CIFAR-10. It is based on a public Lasagne implementation \footnote{\url{https://github.com/Lasagne/Recipes/blob/master/papers/deep_residual_learning/Deep_Residual_Learning_CIFAR-10.py}}, containing $32$ layers. The mini-batch size is set as $M=128$ and Momentum-SGD \cite{momentum_sgd} is used as the optimization algorithm. Following the learning rate scheduling strategy in the original paper \citep{ResNet}, we set the initial learning rate as $0.1$ and multiply it by a factor of $0.1$ after the $32k$-th and $48k$-th model update. Training in this way the test accuracy reaches about $92.4\%$. 

For training NDF,  the sampled training data $D'$ contains $45k$ images, among which $5k$ randomly selected images act as held-out $D'_{dev}$ to provide the reward signal. The other configurations for NDF training, such as state features, policy network structure and optimization algorithm, are almost the same as those in MNIST experiments, except that we train the policy for 100 episodes since ResNet training on Cifar10 is more computationally expensive.
\begin{figure}[htbp]
\centering
\includegraphics[width=1\columnwidth,height=5.4cm]{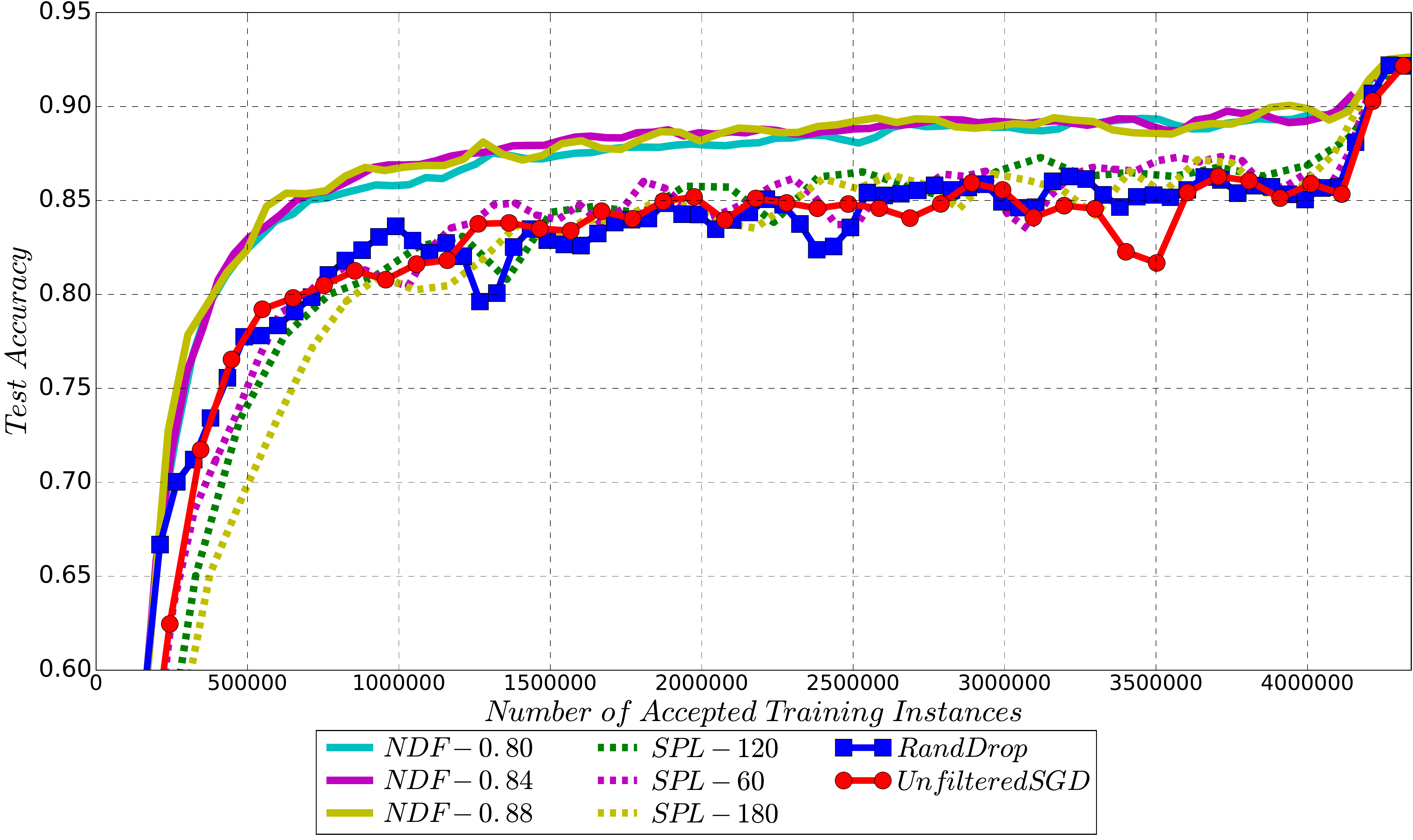}
\caption{Test accuracy curves of training ResNet on Cifar-10 with different data filtration policies. NDF hyper-parameter $\tau\in\{0.80,0.84,0.88\}$, SPL hyper-parameter $S\in\{60,120,180\}$. RandDrop uses the filtered data information output by NDF-$0.84$. }
\label{fig:cifar}
\vspace{-3mm}
\end{figure}

Figure \ref{fig:cifar} records the curves of test accuracy varying with number of effective training data instances, using different data filtration strategies. Once again, NDF outperforms other strategies, as indicated by the fact that to achieve $0.85$ classification accuracy, SGD with NDF spends roughly half training data as that used by SGD without any data selection policy (the Unfiltered-SGD in Figure \ref{fig:cifar}).  In addition, SGD with SPL performs almost the same as SGD with no data filtered, with a tiny gain after training with $3.5m$ instances. However,  
SPL cannot catch up with NDF.

\begin{figure}[htbp]
\centering
\includegraphics[width=1\columnwidth]{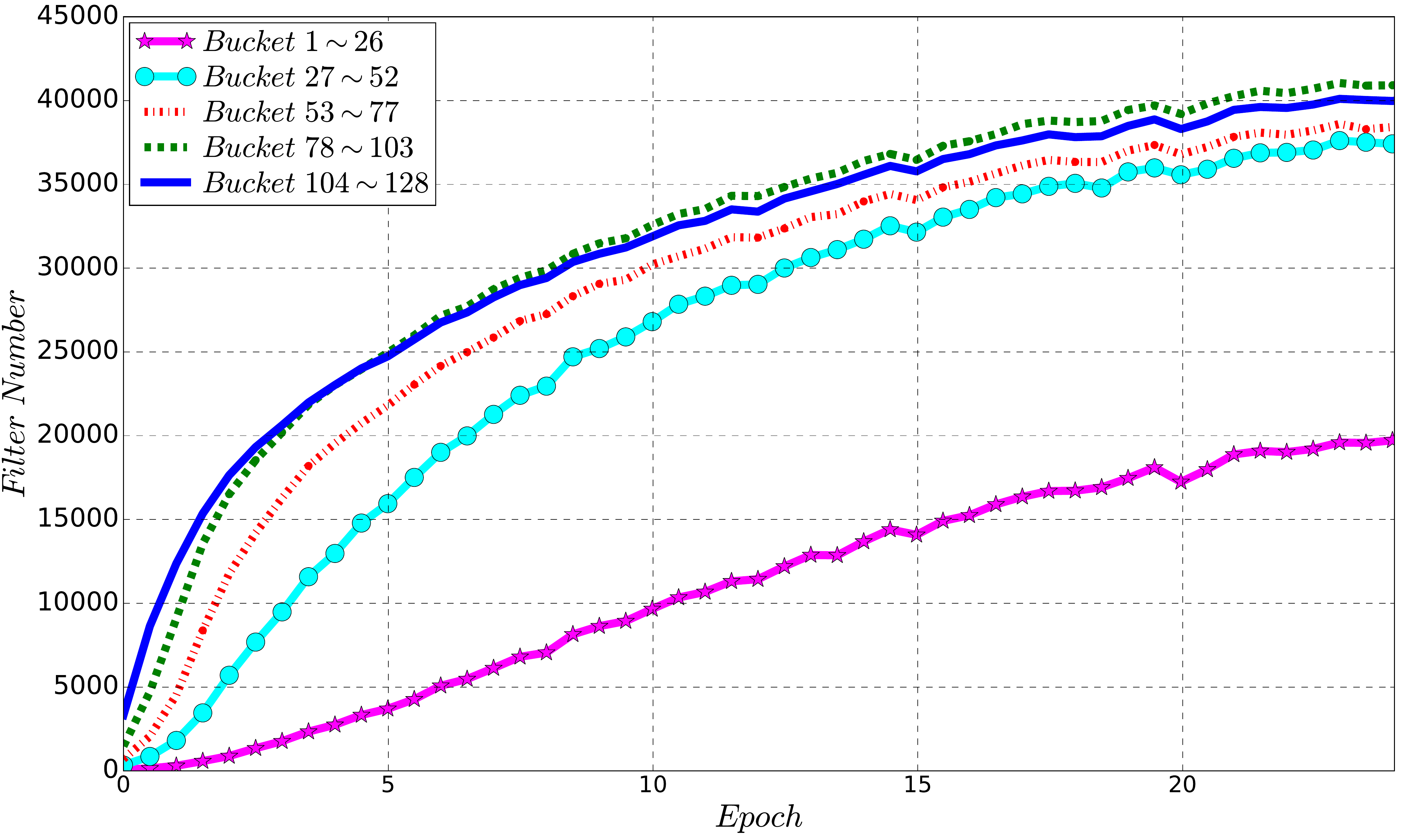}
\caption{The number of instances filtered by NDF in each epoch of Cifar10 training. Similar to Figure \ref{fig:mnist_dp}, we separate ranks $\{1,2,\cdots, 128\}$ of loss values into five buckets to denote training data with different \emph{hardness} levels. } 
\label{fig:cifar_dp}
\vspace{-3mm}
\end{figure}

Similar to Figure \ref{fig:mnist_dp}, we plot the number of instances  filtered by NDF varying with training epochs and different \emph{hardness}  levels in Figure \ref{fig:cifar_dp}.  One can clearly observe a similar data filtration pattern with that of MNIST, yet  quite different with that of SPL, since more and more training data is filtered during learning process and \emph{hard} data instances tend to be kept (shown by the purple line).

\subsection{RNN for IMDB sentiment classification}
In addition to image recognition task, we also test those data selection mechanisms in text related tasks. Here the basic setting is using Recurrent Neural Network (RNN) to conduct sentiment classification. \emph{IMDB movie review dataset}\footnote{\url{http://ai.stanford.edu/~amaas/data/sentiment/}} is a binary sentiment classification dataset consisting of $50k$ movie review comments with positive/negative sentiment labels ~\citep{IMDB}, which are evenly separated (i.e., $25k$/$25k$) as train/test set. The sentences in IMDB dataset are significantly long, with average word token number as $281$. Top $10k$ most frequent words are selected as the dictionary while the others are replaced with a special token UNK.  We apply LSTM ~\citep{LSTM} RNN to each sentence, taking randomly initialized word embedding vectors as input, and the last hidden state of LSTM is fed into a logistic regression classifier to predict the sentiment label ~\citep{semi_s2s}. The size of word embedding in RNN is $256$, the size of hidden state of RNN is $512$, and the mini-batch size is set as $M=16$. Adadelta ~\citep{Adadelta} is used to perform LSTM model training. The test accuracy is roughly $88.5\%$, reproducing the result in previous work~\citep{semi_s2s}.

For NDF training, from all the training data we randomly sample $20k$ as $D'_{train}$ and $3k$ as $D'_{dev}$ to learn data filtration policy. The episode number is set as $L=200$. Early stop on validation set is used to control training process in each episode. The other configurations repeat those used in policy network training in MNIST.
\begin{figure}
\centering
\includegraphics[width=\columnwidth]{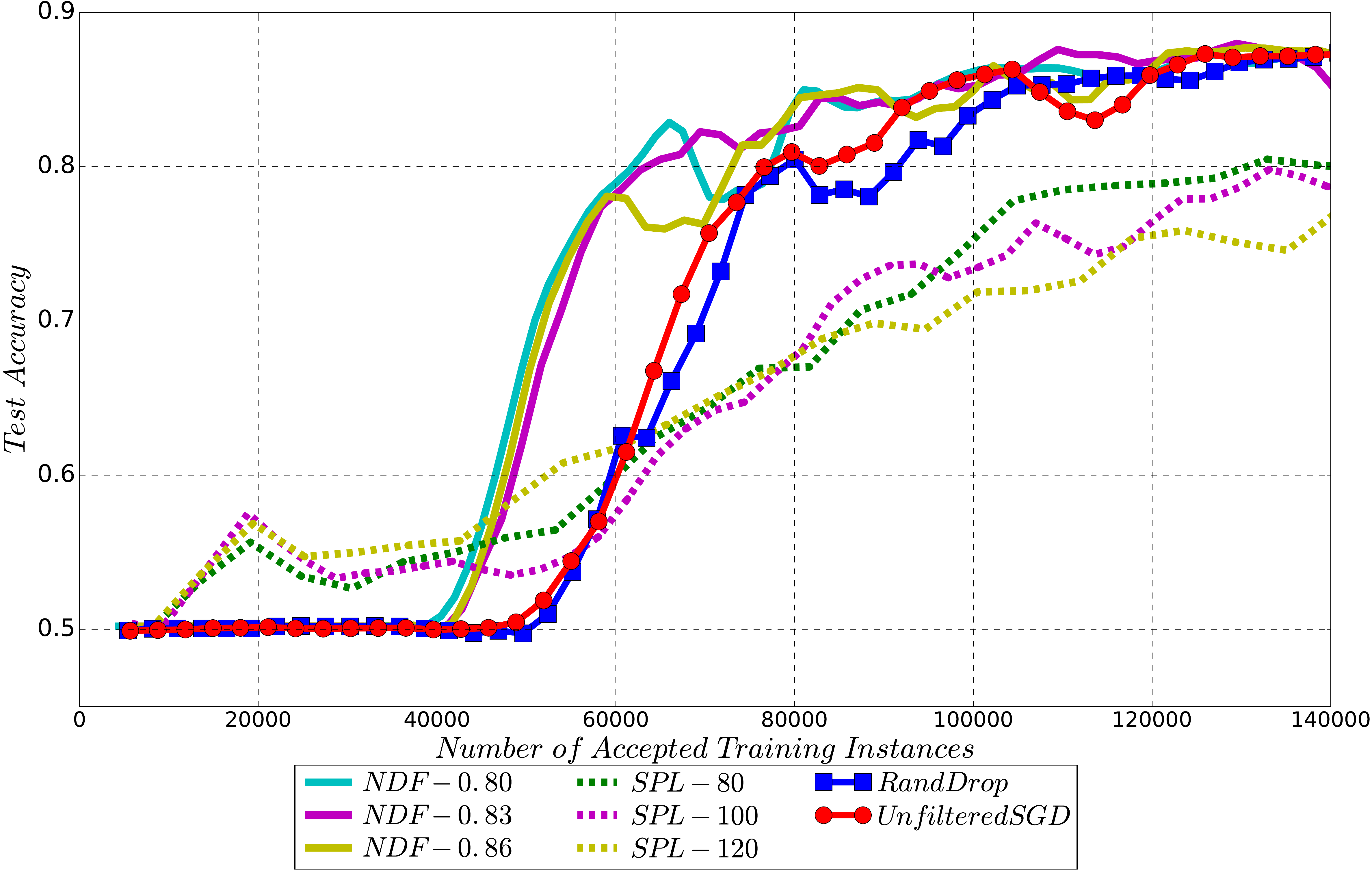}
\caption{Test accuracy curves of different data filtration strategies on IMDB sentiment classification dataset. NDF hyper-parameter $\tau\in\{0.80,0.83,0.86\}$, SPL hyper-parameter $S\in\{80,100,120\}$. RandDrop uses the filtered data information output by NDF with $\tau=0.80$. }
\label{fig:IMDB}
\vspace{-3mm}
\end{figure}

The detailed results are shown in Figure ~\ref{fig:IMDB}.  From the figure we have the following observations:  First,  NDF significantly boosts the convergence of SGD training for LSTM. With much less data, NDF achieves satisfactory classification accuracy. For example, to achieve $80\%$ test accuracy, NDF needs about $75\%$ training data ($60k$) as that of plain Adadelta (about $80k$).  Second, NDF significantly outperforms the RandDrop baseline, demonstrating the effectiveness of learnt policies. At last, self-paced learning (shown by the dashed line) helps for the initialization of LSTM. However, it seems not to help training after the middle phase. Using reinforcement learning,  NDF achieves both better long-term convergence and faster initialization, although the initialization is not as effective as SPL.  


\begin{figure}
\centering
\includegraphics[width=1\columnwidth,height=4.3cm]{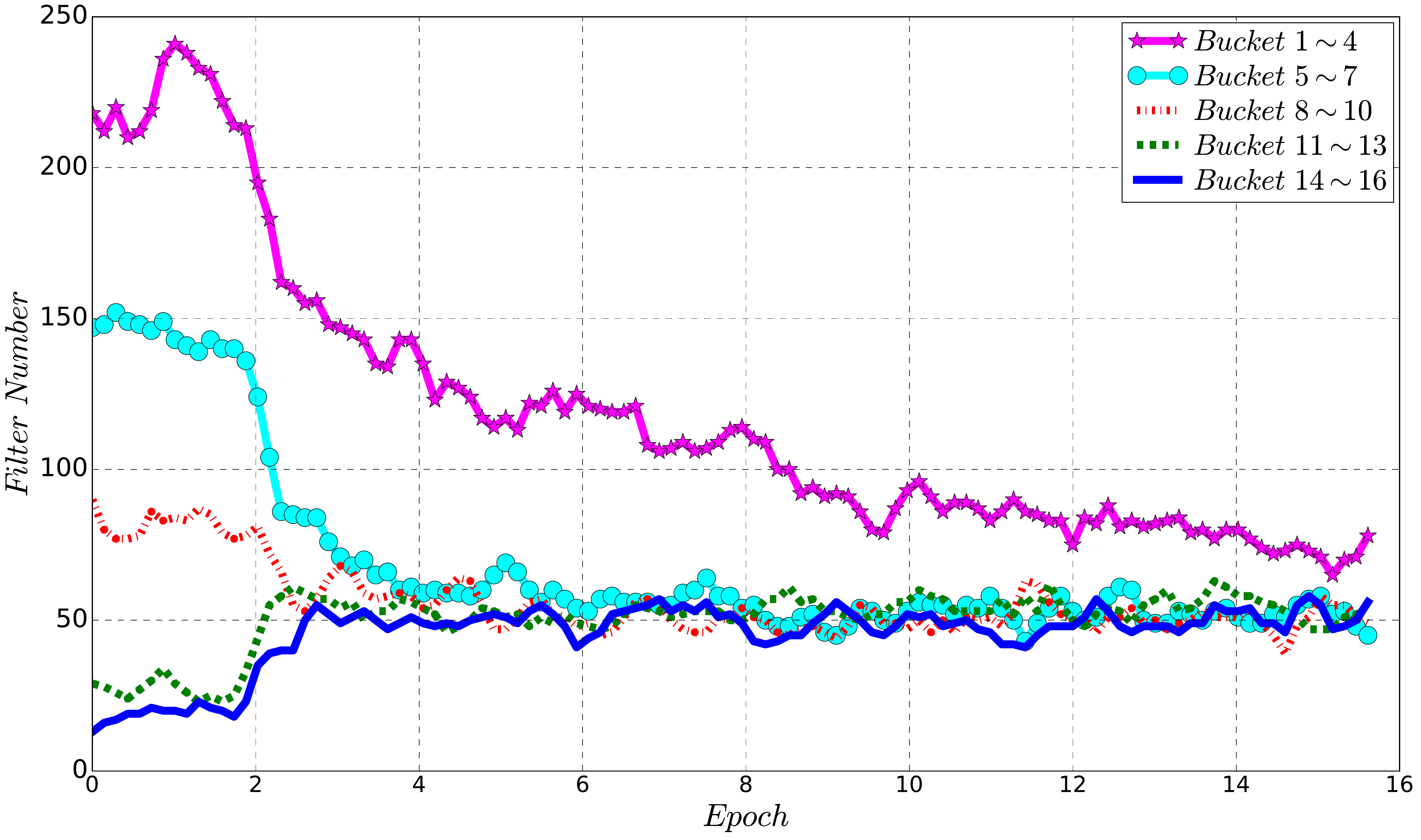}
\caption{The filtered data numbers by NDF in each epoch of IMDB training. The batch size is $16$. } 
\label{fig:imdb_dp}
\vspace{-3mm}
\end{figure}

To better understand the advantages brought of NDF-REINFOFCE, in Figure~\ref{fig:imdb_dp} we also give the curves recording number of filtered data instances in each epoch. One can observe that the data selection pattern learnt by NDF is significantly different with that in MNIST and CIFAR. Particularly, the learnt data selection mechanism is similar with SPL, since \emph{hard} data with larger loss values (by the purple lines) are probably filtered at the early age of LSTM training, while gradually included into model update along the training process.  As shown by the better initialization brought by NDF and SPL compared with no data selection, training with only \emph{easy} data in the early phase helps accelerating LSTM initialization, which has been identified as a difficult problem for LSTM training with long sequences ~\cite{semi_s2s,ResidualRNN}.  From this point of view, it is necessary to start learning from \emph{easy} data. Once the model has grown into a \emph{mature} state with stronger capability of handling various inputs, the \emph{hard} examples will be gradually included. 

However, from the global point of view for Figure~\ref{fig:imdb_dp}, training LSTM on IMDB data favors \emph{easier} data. Except for the aforementioned difficulty of LSTM training, we conjecture another reason for such behavior is compared with image data, natural language contains more noise residing in both the sentences and their labels. Data with large loss values might imply high noise levels, which should be eliminated in model training.


\subsection{Discussion}

We have the following discussions on the experimental results reported above.

\begin{itemize}
\item A good data selection mechanism can effectively accelerate model convergence. For example, the data selection policy learnt by NDF can help the SGD training of various neural networks to achieve fairly good performances, with significantly smaller number of training data than SGD without data section.
\item Different tasks and datasets may favor different data selection policies, as indicated by Figures \ref{fig:mnist_dp}, \ref{fig:cifar_dp} and \ref{fig:imdb_dp}. In this sense, a heuristic data selection rule, such as SPL, cannot cover all different scenarios. In contrast, NDF acts in a more adaptive way and can successfully handle diverse scenarios. This is because it covers a lot of information in its state features that can indicate the importance of a data instance, and it obtains the target data selection policy based on reinforcement learning.
\item Furthermore NDF is not sensitive to the setting of hyper-parameters. Policies trained with a wide range of $\tau$ values can all lead to satisfactory performances.
\item SPL typically works for shallow model training ~\citep{SPL-visual,SPL4MM,SPLDiversity} that does not involve frequent model updated. However, when training deep models with SGD,  SPL does not provide good data selection mechanism with its simple and heuristic rule.
\end{itemize}

\section{Related Work}
\label{sec:Related}
Plenty of previous works talk about data scheduling (e.g., filtration and ordering) strategies for machine learning. A remarkable example is curriculum learning (CL) ~\citep{CL} showing that a data order from \emph{easy} instances to \emph{hard} ones, a.k.a., a \emph{curriculum}, benefits learning process. The measure of \emph{hardness} in CL is typically determined by heuristic understandings of data ~\citep{CL,CL4DependencyParsing,CL4wordembedding}. As a comparison, self-paced learning (SPL) ~\citep{SPL,SPL-visual,SPL4MM,SPLDiversity,SPL4Tracking} quantifies the \emph{hardness} by the loss on data. In SPL, those training instances with loss values larger than a threshold $\eta$ will be neglected and $\eta$ gradually increases in the training process such that finally all training instances will play effects. Apparently SPL can be viewed as a data filtration strategy considered in this paper.

Recently with the revival of deep neural networks, researchers have noticed the importance of data scheduling for deep learning. For example, in ~\citep{OnlineBatchSelection}, a simple batch selection strategy based on the loss values of training data is proposed for speeding up neural network training. ~\citep{CL4wordembedding} leverages bayesian optimization to optimize a curriculum function for training distributed word representations. \citeauthor{CL4QA} (\citeyear{CL4QA}) investigate several hand-crafted criteria for data ordering in solving Question Answering tasks based on DNN. In computer vision, a \emph{hard} example mining approach tailored for training object detection network is proposed in ~\citep{RGB}. Our work differs significantly with these works in that 1) We filter data in randomly arrived mini-batches in training process to save computational efforts, rather than actively select mini-batch through a feedforward process on all the un-trained data, which is quite computationally expensive; 2) We leverage reinforcement learning to automatically derive the optimal data selection policy according to the feedback of training process, rather than use naive and heuristic rules for each task. The latter one is limited and time-consuming,  as show by an example that the complicated rules in~\citep{OnlineBatchSelection} accelerate MNIST training, but fail on Cifar10. In that sense, NDF belongs to the category of meta learning ~\citep{metaLearning, schmidhuber1993neural}, a.k.a., learning to learn ~\citep{L2LBook, L2O,L2R}.

The proposed Neural Data Filter (NDL) for data filtration is based on deep reinforcement learning (DRL) ~\citep{DRLAtari,DRLasynchronous,AlphaGo}, which applies deep neural networks to reinforcement learning ~\citep{RLSutton}. In particular, NDL belongs to policy based reinforcement learning, seeking to search directly for optimal control policy. REINFORCE ~\citep{REINFORCE} and actor-critic~\citep{ActorCritic} are two representative policy gradient algorithms, with the difference that actor-critic adopts value function approximation to reduce the high variance of policy gradient estimator in REINFORCE.

\section{Conclusion}
\label{sec:Conclusion}

In this paper, we have introduced Neural Data Filter (NDF), a reinforcement learning framework to adaptively perform training data selection for machine learning. Experiments on several deep neural networks training by mini-batch SGD have demonstrated that NDF boosts the convergence of training process. On one hand, we have shown that such reinforcement learning based adaptive approach is effective and general for various machine learning tasks; on the other hand, we would like to inspire the community to explore more on data selection/scheduling for machine learning, especially for training deep neural networks.

As to future work, our first goal is to provide more efficient way for NDF training, for example by optimality tightening~\cite{OT}, by Actor-Critic to collect more frequent reward signals\footnote {Our preliminary experiments has verified its effectiveness on IMDB dataset.}, or by learning from pure scratch to eliminate the feed-forward step in current design to obtain the state features.  We further aim to apply such a reinforcement learning based \emph{teacher}-\emph{student} framework to other strategy design problems for machine learning, such as hyper-parameter tuning, structure learning and distributed scheduling, with the hope of providing better guidance for controlled training process.



\nocite{langley00}

\bibliography{NDF_full}
\bibliographystyle{icml2017}

\end{document}